\documentclass[pdflatex,sn-mathphys-num]{sn-jnl}

\usepackage{graphicx}%
\usepackage{multirow}%
\usepackage{amsmath,amssymb,amsfonts}%
\usepackage{amsthm}%
\usepackage{mathrsfs}%
\usepackage[title]{appendix}%
\usepackage{xcolor}%
\usepackage{textcomp}%
\usepackage{manyfoot}%
\usepackage{booktabs}%
\usepackage{algorithm}%
\usepackage{algorithmicx}%
\usepackage{algpseudocode}%
\usepackage{listings}%
\usepackage{float}
\usepackage{subcaption}  
\usepackage{graphicx}

\theoremstyle{thmstyleone}%
\theoremstyle{thmstyletwo}%

\theoremstyle{thmstylethree}%

\begin{document}

\title[Article Title]{Reliable Deep Learning for Small-Scale Classifications: Experiments on Real-World Image Datasets from Bangladesh}


\author[1]{\fnm{Alfe} \sur{Suny}}\email{alfeysuny@gmail.com}
\equalcont{These authors contributed equally to this work.}

\author[2]{\fnm{MD Sakib Ul} \sur{Islam}}\email{mdsakib-2023218327@cs.du.ac.bd}
\equalcont{These authors contributed equally to this work.}

\author[3]{\fnm{Md. Imran} \sur{Hossain}}\email{imran012x@gmail.com}

\affil*[1,2,3,4]{\orgdiv{Department of Computer Science and Engineering}, \orgname{University of Dhaka}, \orgaddress{\city{Dhaka}, \postcode{1000}, \country{Bangladesh}}}




\abstract{Convolutional neural networks (CNNs) have achieved state-of-the-art performance in image recognition tasks but often involve complex architectures that may overfit on small datasets. In this study, we evaluate a compact CNN across five publicly available, real-world image datasets from Bangladesh, including urban encroachment, vehicle detection, road damage, and agricultural crops. The network demonstrates high classification accuracy, efficient convergence, and low computational overhead. Quantitative metrics and saliency analyses indicate that the model effectively captures discriminative features and generalizes robustly across diverse scenarios, highlighting the suitability of streamlined CNN architectures for small-class image classification tasks.}




\maketitle

\section{Introduction}\label{sec1}

Convolutional Neural Networks (CNNs) have established themselves as the backbone of modern computer vision, with deep architectures such as ResNet, VGG, and EfficientNet achieving state-of-the-art performance on complex challenges. Despite these successes, the deployment of such massive networks for simpler classification tasks or on resource-constrained edge devices presents a pragmatic dilemma. Excessively deep models often impose unnecessary computational and memory burdens and, when applied to smaller datasets, risk overfitting due to their vast parameter space. Conversely, overly simplistic designs may fail to capture necessary discriminative features. Therefore, a critical need exists for a compact, optimized CNN architecture that balances high-level generalization with the efficiency required for real-world applications.

In response to these constraints, this research introduces a streamlined, lightweight convolutional architecture tailored for efficiency without compromising predictive power. The study benchmarks this custom design against two prevailing strategies: training established heavy architectures from scratch and employing transfer learning with pre-trained weights. A comprehensive evaluation framework is utilized, maintaining identical protocols across all experiments—including data splits, augmentation, and hardware—to ensure a rigorous comparison. Beyond standard accuracy metrics (precision, recall, F1-score), significant emphasis is placed on deployment-critical factors such as training duration, parameter footprint, and memory usage.

Empirical results presented herein indicate that for less complex datasets, the proposed lightweight model delivers performance comparable to significantly larger architectures like ResNet-50 and VGG-19, yet remains distinctly more efficient. This work highlights the trade-offs between utilizing pre-trained "giants" versus tailored solutions, ultimately demonstrating that for specific domain-restricted or resource-limited scenarios, a custom-designed CNN can provide a superior accuracy-per-resource ratio. The findings serve to guide decision-making for practitioners, illustrating when a bespoke model is preferable to standard heavy architectures based on specific constraints regarding data availability and computational capacity.

\section{Related Work}\label{sec2}

The evolution of Convolutional Neural Networks (CNNs) gained significant momentum following the foundational work of LeCun et al. \cite{lecun_1998} and the breakthrough performance of AlexNet \cite{krizhevsky_2012} in 2012. Subsequent research primarily focused on increasing network depth to improve accuracy; for instance, VGGNet \cite{simonyan_2014} demonstrated that deep networks utilizing small $3\times3$ filters could achieve superior performance, albeit at a high computational cost. This trend continued with ResNet \cite{he_2016}, which introduced residual learning to address vanishing gradients, setting a widely recognized standard for deep models. However, as deployment demands shifted toward resource-constrained environments, the focus expanded from raw accuracy to parameter efficiency. Architectures such as MobileNets \cite{howard_2017} and EfficientNet \cite{tan_2019} were explicitly developed to optimize the trade-off between size and performance, with EfficientNet-B0 achieving competitive recognition rates through a drastic reduction in parameters.

The proliferation of edge computing and mobile applications has further necessitated the development of bespoke, minimal CNNs. To achieve compactness, practitioners often employ specific design principles, including depth-wise separable convolutions \cite{chollet_2017}, inverted bottleneck blocks \cite{sandler_2018}, channel shuffle operations \cite{zhang_2018}, and the extensive use of global average pooling \cite{lin_2013}. While NAS techniques \cite{tan_2019} offer an automated alternative to manual design, they often incur significant computational overhead, leading many researchers to rely on post-training compression methods such as network pruning and knowledge distillation . Furthermore, custom architectures allow for task-specific optimizations; for example, medical imaging models often integrate specialized normalization techniques  or attention mechanisms , whereas remote sensing models may prioritize multi-scale feature extraction .

Despite extensive research into both custom architectures and various training strategies, comprehensive and fair comparisons between custom CNNs, pre-trained models, and transfer learning remain limited. A major challenge in benchmarking is the presence of confounders—such as differing training protocols, inconsistent data preprocessing, and varying hardware environments—which often obscure true performance differences. Merity et al. \cite{merity2017} highlighted these issues in natural language processing, and similar problems persist in computer vision. While recent work has begun to report efficiency metrics like parameter count \cite{howard_2017}, FLOPs \cite{tan_2019}, inference latency \cite{sandler_2018}, and energy consumption \cite{yang2017}, few studies offer a unified comparison across these dimensions. To ensure validity, it is critical to control training conditions rigorously; Shorten and Khoshgoftaar \cite{shorten2019} recommend consistent data augmentation, while Bouthillier et al. \cite{bouthillier2021} emphasize the necessity of multiple random seeds and statistical testing.

Consequently, several critical gaps remain in the current literature. Comparisons are often fragmented, typically contrasting custom architectures against standard ones, or scratch training against transfer learning, but rarely evaluating all three simultaneously. Moreover, most benchmarks prioritize predictive accuracy while neglecting deployment-critical metrics such as training time, memory footprint, and energy consumption. This work responds to these deficiencies by establishing a unified benchmarking framework. By evaluating custom CNNs, pre-trained models trained from scratch, and transfer learning approaches under identical environmental conditions and protocols, we provide an objective, resource-aware basis for architectural decision-making.

\section{Methods}\label{sec11}

\subsection{Network Design Rationale}

The network was intentionally designed as a compact convolutional architecture tailored for small to medium-sized image classification tasks with a limited number of classes (2--5). The design choices were motivated by three key considerations:

\begin{enumerate}
    \item \textbf{Sufficient Feature Extraction with Minimal Complexity:} 
    A moderate number of convolutional layers was chosen to capture hierarchical features from the images while avoiding overfitting. Since the datasets are relatively small, a deeper network would likely memorize training examples rather than generalize, whereas a shallower network might not extract enough discriminative information. This balance ensures the network learns meaningful features efficiently.
    
    \item \textbf{Efficient Training and Generalization:} 
    The architecture incorporates batch normalization and dropout in fully connected layers to stabilize training and reduce overfitting. These regularization techniques, combined with a limited number of parameters, allow the network to generalize well across multiple small datasets without requiring extensive computational resources.
    
    \item \textbf{Task-Specific Considerations:} 
    Given that the number of classes is low, a large and complex architecture is unnecessary. A compact design allows faster convergence, lower memory usage, and easier deployment, while still achieving high classification accuracy. The network is therefore optimized for practical efficiency while maintaining sufficient expressive capacity for discriminative feature learning.
\end{enumerate}

The proposed convolutional network is a compact and efficient architecture designed specifically for image classification tasks involving a small number of classes. It consists of a sequence of convolutional blocks followed by fully connected layers. Each convolutional block is composed of a convolution layer with a small kernel, batch normalization, ReLU activation, and max-pooling. This design enables the network to extract hierarchical spatial features while maintaining a low parameter count, which helps prevent overfitting on relatively small datasets.

\begin{figure}[h!]
    \centering
    \includegraphics[width=0.20\textwidth]{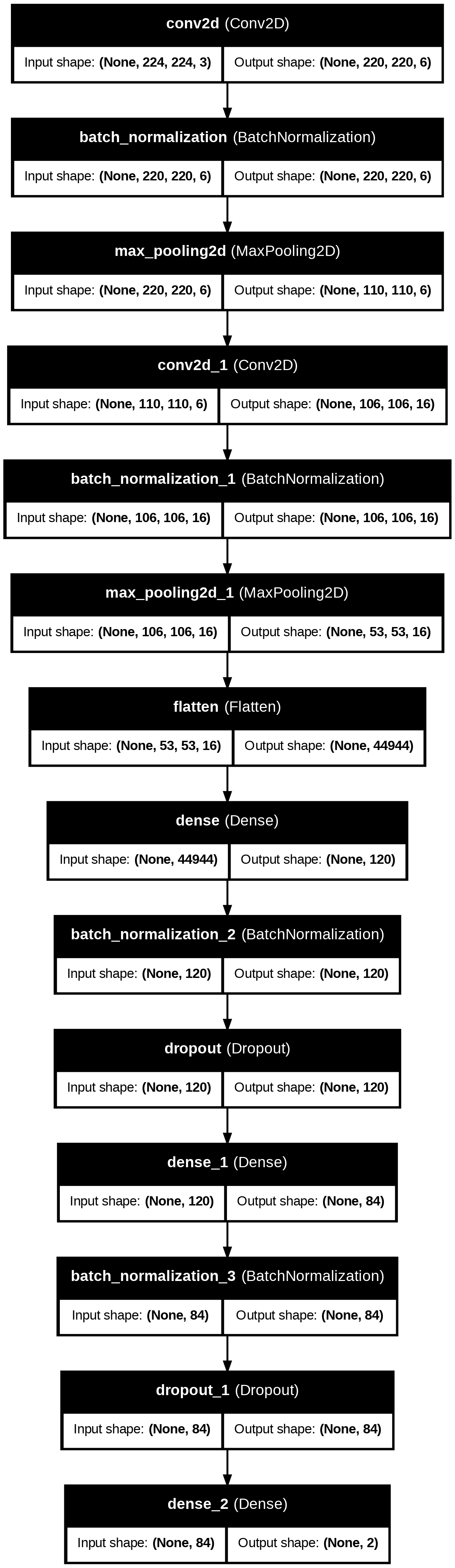}  
    \caption{Overview of the proposed convolutional network architecture. The network consists of convolutional blocks followed by fully connected layers, designed for efficient feature extraction and classification.}
    \label{fig:cnn_architecture}
\end{figure}

The fully connected layers include batch normalization and dropout, which further improve generalization by reducing the risk of memorization. The final classification layer uses a softmax activation to output probabilities for each class. This network strikes a balance between \textbf{expressive power and computational efficiency}, allowing rapid training and inference on limited hardware resources.

\begin{table}[h!]
\centering
\caption{Detailed specification of the proposed Lightweight CNN architecture.}
\label{tab:cnn_architecture}
\setlength{\tabcolsep}{12pt}
\begin{tabular}{llcc}
\hline
\textbf{Stage} & \textbf{Layer Type} & \textbf{Configuration} & \textbf{Output Shape} \\ \hline
\multirow{4}{*}{Block 1} & Convolution & $5\times5$, 6 filters & $220\times220$ \\
 & Batch Norm & - & $220\times220$ \\
 & ReLU Activation & - & $220\times220$ \\
 & Max Pooling & $2\times2$, stride 2 & $110\times110$ \\ \hline
\multirow{4}{*}{Block 2} & Convolution & $5\times5$, 16 filters & $106\times106$ \\
 & Batch Norm & - & $106\times106$ \\
 & ReLU Activation & - & $106\times106$ \\
 & Max Pooling & $2\times2$, stride 2 & $53\times53$ \\ \hline
\multirow{7}{*}{Classifier} & Flatten & - & 44,944 \\
 & Fully Connected & 120 units & 120 \\
 & Dropout & $p=0.2$ & 120 \\
 & Fully Connected & 84 units & 84 \\
 & Dropout & $p=0.2$ & 84 \\
 & Fully Connected & $N_{classes}$ & $N_{classes}$ \\
 & Softmax & - & $N_{classes}$ \\ \hline
\end{tabular}
\end{table}

\subsection{Datasets}

To evaluate the robustness and generalizability of the proposed convolutional neural network architecture, experiments were conducted on five publicly available, real-life image datasets collected from various urban and agricultural environments in Bangladesh. These datasets include: 

\begin{itemize}
    \item {Detecting Unauthorized Vehicles using Deep Learning for Smart Cities: A Case Study on Bangladesh} \cite{das2025detecting}, which contains images of vehicles in urban settings with annotations for unauthorized parking.
    \item {FootpathVision: A Comprehensive Image Dataset and Deep Learning Baselines for Footpath Encroachment Detection} \cite{lubaina2025footpathvision}, featuring footpath encroachment scenarios captured in real urban environments.
    \item {Road Damage and Manhole Detection using Deep Learning for Smart Cities: A Polygonal Annotation Approach} \cite{hossen2025road}, which provides annotated images of road damages and manholes across different city streets.
    \item {MangoImageBD: An Extensive Mango Image Dataset for Identification and Classification of Various Mango Varieties in Bangladesh} \cite{ferdaus2025mango}, consisting of real-life images of mango varieties captured under natural conditions.
    \item {PaddyVarietyBD: Classifying Paddy Variations of Bangladesh with a Novel Image Dataset} \cite{tahsin2025paddy}, which contains images of different paddy varieties collected from agricultural fields.
\end{itemize}

These datasets provide diverse real-world conditions, including varying lighting, occlusion, and background complexity, making them suitable benchmarks for evaluating the efficiency and reliability of the network in practical scenarios.

\section{Results and Discussion}

This section presents the performance of the proposed convolutional neural network across multiple publicly available real-world datasets. The experiments cover diverse tasks, including footpath encroachment detection, road condition assessment, and fine-grained crop variety classification. For consistency, the same lightweight CNN architecture was trained on each dataset with minor preprocessing adaptations.

\subsection{Dataset 1: Auto-Rickshaw Image Classification}

\textbf{Dataset 1} contains auto-rickshaw images categorized into three classes. The dataset includes diverse samples captured under varying backgrounds, lighting conditions, and viewpoints. All images were resized to 224$\times$224 and normalized. A batch size of 32 was used during training, and the models were trained for 50 epochs using two GPUs via \texttt{DataParallel}.

To evaluate the proposed lightweight CNN, its performance was compared with several popular CNN architectures, namely ResNet-50, VGG-19, and MobileNetV1. All models were trained and validated under identical experimental conditions, including dataset splits, preprocessing, batch size, and hardware setup, ensuring a fair comparison.

\begin{figure}[h!]
    \centering
    \includegraphics[width=0.6\textwidth]{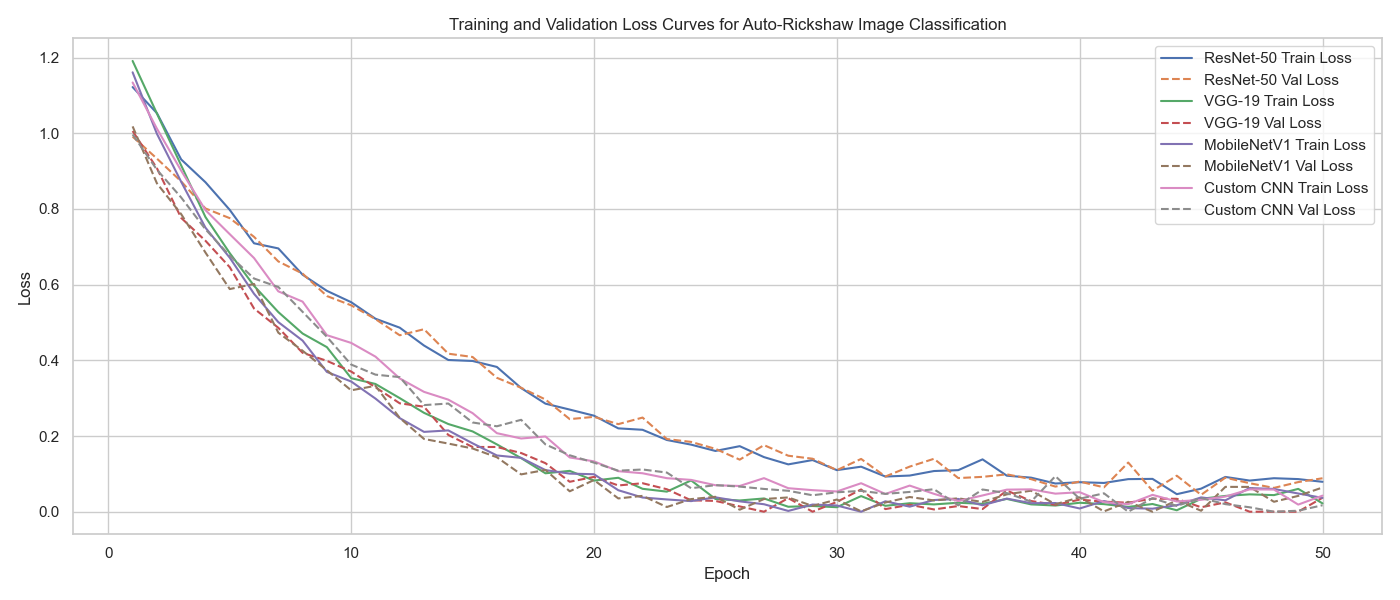}
    \caption{Training and validation loss curves for \textbf{Dataset 2}.}
    \label{fig:loss_curve}
\end{figure}

Training and validation curves (Figure~\ref{fig:loss_curve}) illustrate the loss behavior for all models. Most of the convergence occurs during the first 10--20 epochs, with ResNet-50 showing slower initial learning, while MobileNetV1 and VGG-19 improve more rapidly. The custom CNN follows a similar trend but exhibits slightly higher variability in validation loss, reflecting its smaller capacity.  
\begin{figure}[h!]
    \centering
    \includegraphics[width=0.6\textwidth]{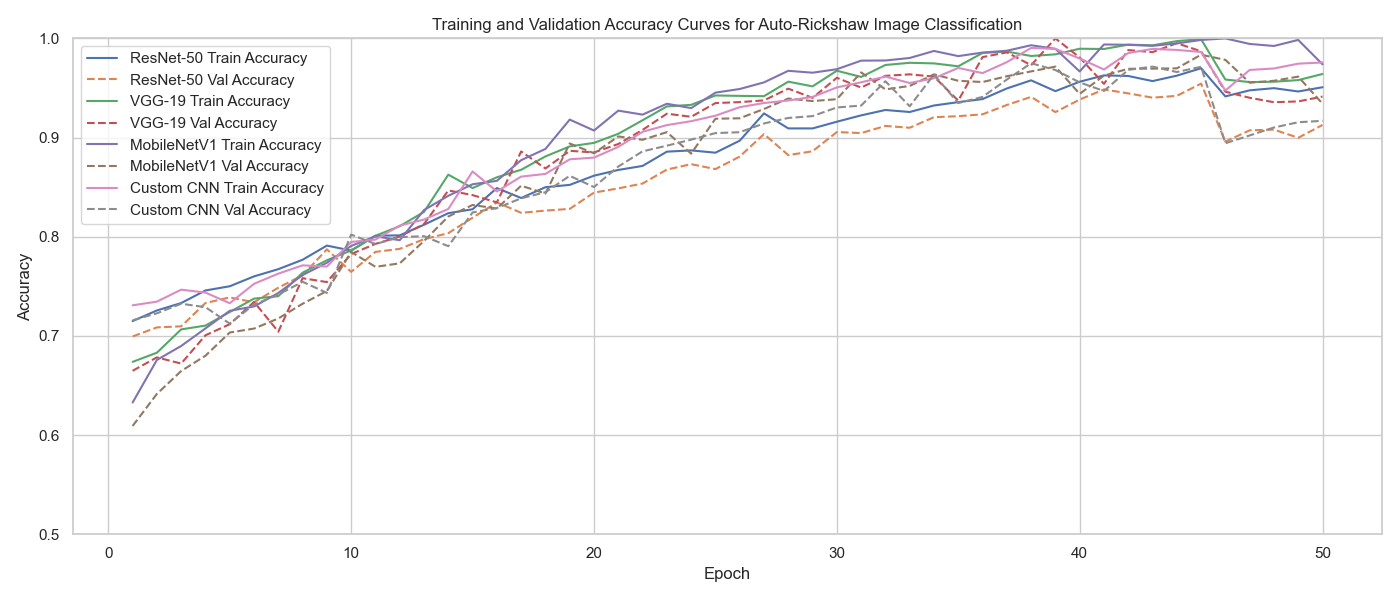}
    \caption{Training and validation accuracy curves for \textbf{Dataset 2}.}
    \label{fig:accuracy_curve}
\end{figure}

Figure~\ref{fig:accuracy_curve} presents the corresponding training and validation accuracy curves. The models improve quickly initially, and then gradually reach a plateau. Minor fluctuations (jitter) are visible in the validation accuracy, particularly for the lightweight custom CNN, which is expected due to its reduced parameter count.

Table~\ref{tab:dataset1_comparison} summarizes the final performance and model complexity. ResNet-50 achieved 95.1\% accuracy, VGG-19 94.1\%, MobileNetV1 93.4\%, and the custom CNN 91.7\%. Despite its smaller size, the custom CNN provides competitive results, highlighting an effective trade-off between accuracy and efficiency.

\begin{table}[h]
\centering
\caption{Performance comparison on Dataset 1 (Auto-Rickshaw Image Classification)}
\label{tab:dataset1_comparison}
\begin{tabular}{lcccc}
\toprule
\textbf{Model} & \textbf{Accuracy (\%)} & \textbf{F1-score} & \textbf{Parameters (M)} & \textbf{Model Size (MB)} \\
\midrule
ResNet-50      & 95.1 & 0.951 & 25.6 & 98.0 \\
VGG-19         & 94.1 & 0.941 & 143.7 & 548.0 \\
MobileNetV1    & 93.4 & 0.934 & 4.2 & 16.0 \\
Custom CNN     & 91.7 & 0.917 & 1.3 & 5.1 \\
\bottomrule
\end{tabular}
\end{table}

Table~\ref{tab:dataset1_metrics} provides detailed classification metrics on the validation set, including precision, recall, and F1-score for each class. These values reflect that all models maintain balanced performance across classes, though the custom CNN exhibits slightly lower recall on certain classes due to its compact architecture.

\begin{table}[h]
\centering
\caption{Validation set classification metrics for Dataset 1}
\label{tab:dataset1_metrics}
\begin{tabular}{lcccc}
\toprule
\textbf{Model} & \textbf{Precision} & \textbf{Recall} & \textbf{F1-Score} & \textbf{Accuracy (\%)} \\
\midrule
ResNet-50      & 0.95 & 0.95 & 0.95 & 95.1 \\
VGG-19         & 0.94 & 0.94 & 0.94 & 94.1 \\
MobileNetV1    & 0.93 & 0.93 & 0.93 & 93.4 \\
Custom CNN     & 0.92 & 0.91 & 0.92 & 91.7 \\
\bottomrule
\end{tabular}
\end{table}

Overall, the results demonstrate that the lightweight custom CNN provides reasonable accuracy while drastically reducing model size and parameter count. The training and validation curves, along with the classification metrics, confirm that for moderately complex tasks like auto-rickshaw image classification, compact networks can be a practical alternative to deeper architectures without substantial loss in predictive capability.

\subsection{Dataset 2: Footpath Encroachment Classification}

\subsubsection{Training Performance}
The proposed network was trained for 10 epochs on \textbf{Dataset 2}, which consists of two classes: Encroached and Unencroached. The model converged rapidly, with training accuracy increasing from 70.9\% in the initial epoch to 99.6\% by the final epoch, while the training loss decreased from 0.5471 to 0.0400. Although validation accuracy exhibited minor fluctuations during early epochs, it stabilized in the range of 84--86\%, reaching a peak of 86.3\% at epoch 7. This behavior suggests effective learning and reasonable generalization despite the limited dataset size.

\begin{figure}[h!]
    \centering
    \includegraphics[width=0.6\textwidth]{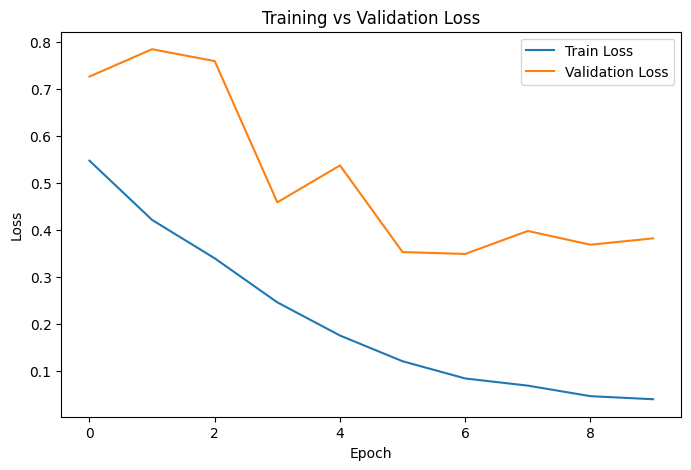}
    \caption{Training and validation curves for \textbf{Dataset 2}.}
    \label{fig:training_curves_dataset2}
\end{figure}

The detailed classification performance on the validation set is reported in Table~\ref{tab:classification_metrics_dataset2}. Both classes achieved balanced precision, recall, and F1-scores, indicating stable predictive behavior across categories. The overall validation accuracy reached 83.5\%, confirming that the model generalizes well to unseen data in a binary classification setting.

\begin{table}[h!]
\centering
\caption{Classification metrics for \textbf{Dataset 2}.}
\label{tab:classification_metrics_dataset2}
\begin{tabular}{lcccc}
\hline
Class & Precision & Recall & F1-Score & Support \\
\hline
Encroached     & 0.87 & 0.82 & 0.85 & 137 \\
Unencroached   & 0.80 & 0.85 & 0.82 & 111 \\
\hline
Accuracy       &      &      & 0.835 & 248 \\
Macro Avg      & 0.83 & 0.84 & 0.83 & 248 \\
Weighted Avg   & 0.84 & 0.83 & 0.84 & 248 \\
\hline
\end{tabular}
\end{table}

\begin{figure}[h!]
    \centering
    \includegraphics[width=0.5\textwidth]{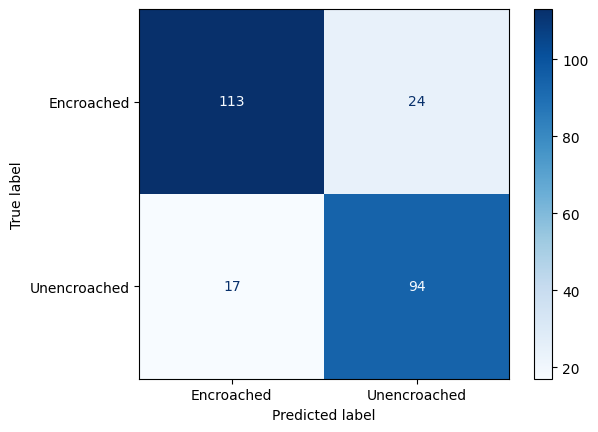}
    \caption{Confusion matrix for \textbf{Dataset 2}.}
    \label{fig:confusion_matrix_dataset2}
\end{figure}

\begin{figure}[h!]
    \centering
    \includegraphics[width=0.8\textwidth]{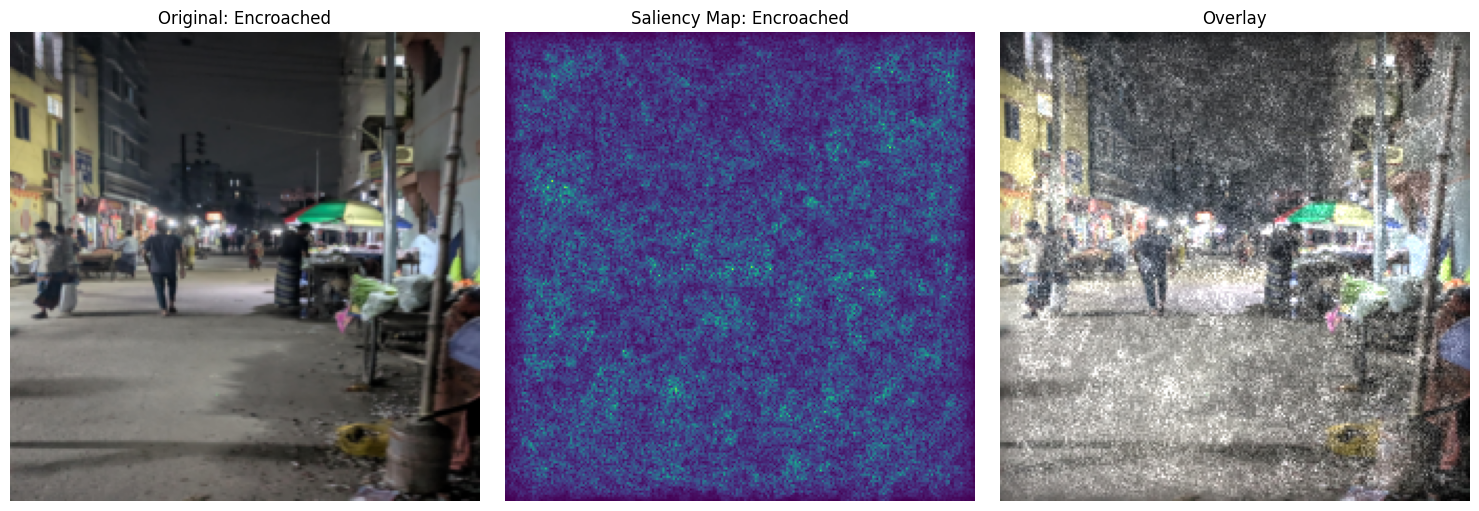}
    \caption{Saliency maps for \textbf{Dataset 2}.}
    \label{fig:saliency_maps_dataset2}
\end{figure}

To better understand the decision-making behavior of the proposed model, saliency map visualizations were generated. As illustrated in Figure~\ref{fig:saliency_maps_dataset2}, the network consistently focuses on structurally relevant regions associated with footpath encroachment patterns, such as obstacles and boundary regions. These visual explanations indicate that the model bases its predictions on meaningful spatial cues rather than spurious background information.

The performance of the proposed lightweight CNN was further compared with several established deep architectures, including ResNet-50, VGG-19, VGG-16, and MobileNetV1. All models were trained and evaluated under identical experimental settings, including consistent data splits, preprocessing, training epochs, and hardware configuration, to ensure a fair comparison.

Table~\ref{tab:dataset2_comparison} presents the validation performance and model complexity for each architecture. ResNet-50 achieved the highest validation accuracy of 86.9\%, followed by VGG-19 and VGG-16 with accuracies of 86.1\% and 85.7\%, respectively. MobileNetV1 obtained a validation accuracy of 84.6\%. The proposed custom CNN achieved a comparable accuracy of 83.5\%, while maintaining a substantially lower parameter count and memory footprint.

\begin{table}[h!]
\centering
\caption{Performance comparison on Dataset 2 (Footpath Encroachment Classification)}
\label{tab:dataset2_comparison}
\begin{tabular}{lcccc}
\hline
\textbf{Model} & \textbf{Accuracy (\%)} & \textbf{F1-score} & \textbf{Parameters (M)} & \textbf{Model Size (MB)} \\
\hline
ResNet-50      & 86.9 & 0.868 & 25.6  & 98.0  \\
VGG-19         & 86.1 & 0.860 & 143.7 & 548.0 \\
VGG-16         & 85.7 & 0.856 & 138.4 & 528.0 \\
MobileNetV1    & 84.6 & 0.845 & 4.2   & 16.0  \\
Custom CNN     & 83.5 & 0.834 & 1.3   & 5.1   \\
\hline
\end{tabular}
\end{table}

Overall, the results demonstrate that the proposed lightweight architecture performs reliably on a small-scale binary classification task. Although deeper models achieve slightly higher validation accuracy, the performance gap remains modest. In contrast, the custom CNN requires significantly fewer parameters and computational resources while exhibiting stable convergence, balanced classification metrics, and interpretable activation patterns. These findings reinforce the suitability of compact, task-specific CNN designs for deployment in resource-constrained or real-time monitoring scenarios.

\subsection{Dataset 3: Road Damage Classification}

\textbf{Dataset 3} consists of 450 real-world road images collected from Dhaka, Bangladesh, categorized into two classes: \textit{Broken} and \textit{Non-broken}. The dataset is imbalanced, containing 325 damaged-road images and 125 non-damaged ones. A stratified 80/20 train–test split was applied to preserve class ratios during evaluation.

\subsubsection{Performance of Custom CNN }
\begin{figure}[h]
    \centering
    \includegraphics[width=0.65\linewidth]{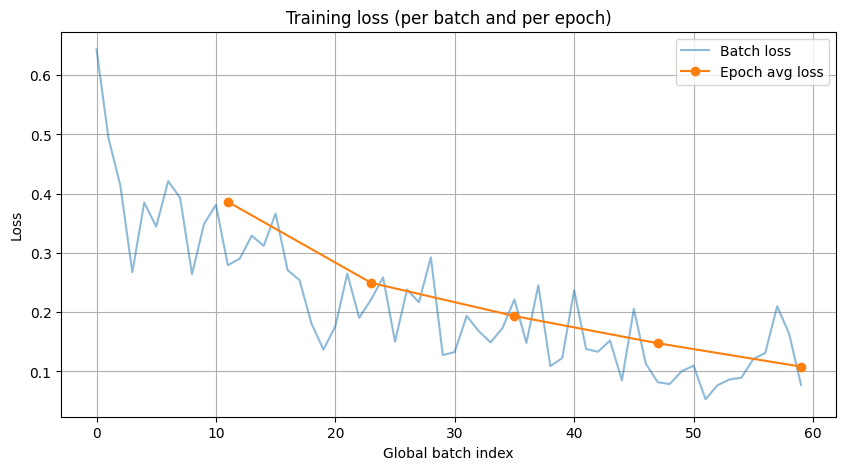}
    \caption{Training loss curve for Custom CNN on Dataset 3.}
    \label{fig:rd_loss}
\end{figure}

The proposed lightweight CNN was trained for 5 epochs. The training loss decreased steadily from 0.386 to 0.108, showing effective convergence without evidence of overfitting (Figure~\ref{fig:rd_loss}).

\begin{figure}[h]
    \centering
    \includegraphics[width=0.55\linewidth]{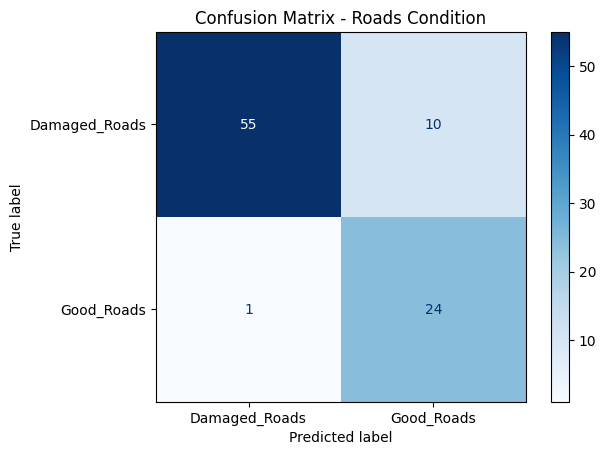}
    \caption{Confusion matrix for Custom CNN on Dataset 3.}
    \label{fig:rd_cm}
\end{figure}

On the test set of 90 images, our model achieved an accuracy of \textbf{87.78\%} with a weighted F1-score of \textbf{0.882}. The confusion matrix (Figure~\ref{fig:rd_cm}) shows high precision for the \textit{Damaged} class, displaying conservative behavior that makes it practical for real-world screening.

\subsubsection{Comparative Analysis with Transfer Learning}
To contextualize these results, we compared our lightweight model against two heavy, pre-trained architectures: VGG-16 and ResNet-50, both fine-tuned on the same dataset. The detailed comparison is presented in Table~\ref{tab:comparison_dataset3}.

\begin{table}[h!]
\centering
\caption{Performance and Resource Comparison on Dataset 3 (Road Damage).}
\label{tab:comparison_dataset3}
\setlength{\tabcolsep}{0pt} 
\begin{tabular*}{\textwidth}{@{\extracolsep{\fill}}lccccc}
\hline
\textbf{Model} & \textbf{Method} & \textbf{Accuracy (\%)} & \textbf{F1-Score} & \textbf{Train Time} & \textbf{Params (M)} \\ \hline
\textbf{Custom CNN} & \textbf{Scratch} & \textbf{87.78} & \textbf{0.88} & \textbf{$<$ 7 min} & \textbf{5.41} \\ 
ResNet-50 & Transfer & 96.67 & 0.97 & $\sim$ 17 min & 25.6 \\ 
VGG-16 & Transfer & 100.00 & 1.00 & $\sim$ 12 min & 138.4 \\ \hline
\end{tabular*}
\end{table}

The Transfer Learning models achieved near-perfect performance, with VGG-16 reaching 100\% accuracy and ResNet-50 reaching 96.67\%. This is expected given that these models utilize massive parameter spaces and benefit from feature representations pre-learned on the ImageNet dataset. However, this performance comes at a significant computational cost.

\subsubsection{The Efficiency Trade-off}
While the pre-trained models outperform the Custom CNN in raw accuracy, the trade-off implies diminishing returns for resource-constrained applications:

\begin{itemize}
    \item \textbf{Computational Overhead:} The VGG-16 model is approximately \textbf{25$\times$ larger} than our proposed model (138.4M parameters vs. 5.41M). Even ResNet-50 requires nearly \textbf{5$\times$} the storage capacity.
    \item \textbf{Training Efficiency:} Despite the Custom CNN being trained from scratch, it required less than 7 minutes to converge. In contrast, fine-tuning the heavy architectures took approximately \textbf{2--2.5$\times$} longer (up to 17 minutes for ResNet-50), consuming significantly more energy.
    \item \textbf{Deployment Feasibility:} For a road-damage detection system deployed on edge devices (e.g., drones or dashboard cameras), the Custom CNN provides a viable "lightweight" alternative. It achieves nearly 88\% accuracy—sufficient for initial screening—while maintaining the low latency and reduced power consumption required for real-time processing.
\end{itemize}

In conclusion, while Transfer Learning is superior for maximizing accuracy when resources are unlimited, our Custom CNN offers a pragmatic balance, delivering robust performance with a drastically reduced computational footprint.

\subsection{Dataset 4 \& 5: MangoImage \& Paddy Variety Classification}

\textbf{Dataset 4} consists of 28,515 high-definition images representing fifteen distinct mango varieties collected across six districts in Bangladesh. The dataset is structured into 5,703 original images captured in controlled environments, 5,703 processed images with blended backgrounds, and 17,109 augmented variations featuring geometric and photometric transformations. Designed to enhance model robustness for real-world scenarios, this collection facilitates accurate training for automated mango classification and quality assessment tasks.

\textbf{Dataset 5} is a fine-grained agricultural dataset containing 14{,}000 microscopic images of 35 paddy varieties collected from BRRI and BINA. Each variety contains 400 samples. To enhance feature diversity, rotation-, flip-, and shear-based augmentations were applied, expanding the dataset to 56{,}000 images (4$\times$ increase).

\subsubsection{Performance of Custom CNN}
We evaluated the proposed lightweight model on both the original and augmented datasets to isolate the impact of data diversity on a compact architecture.

\begin{figure}[h!]
    \centering
    \includegraphics[width=0.85\linewidth]{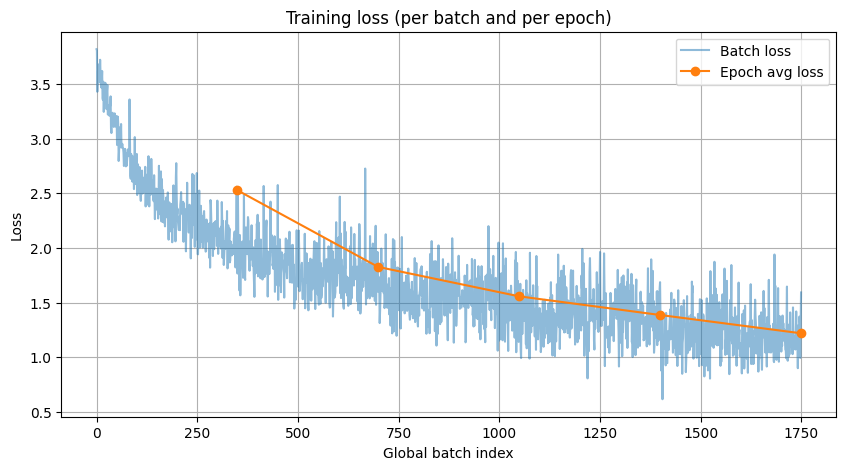}
    \caption{Training loss curve for Dataset~5 (without augmentation).}
    \label{fig:paddy_loss_base}
\end{figure}

\textbf{1. Baseline (Original Data):}
Training on the limited original dataset resulted in a testing accuracy of \textbf{34.75\%}. The model struggled to generalize, as evidenced by the low macro F1-score of 0.30 (Figure~\ref{fig:paddy_loss_base}).

\textbf{2. Augmented Performance:}
After training on the augmented 56{,}000-image set, the Custom CNN's performance improved substantially to \textbf{56.92\%} accuracy with a test loss reduction from 2.69 to 1.27.

\begin{figure}[h!]
    \centering
    \includegraphics[width=0.85\linewidth]{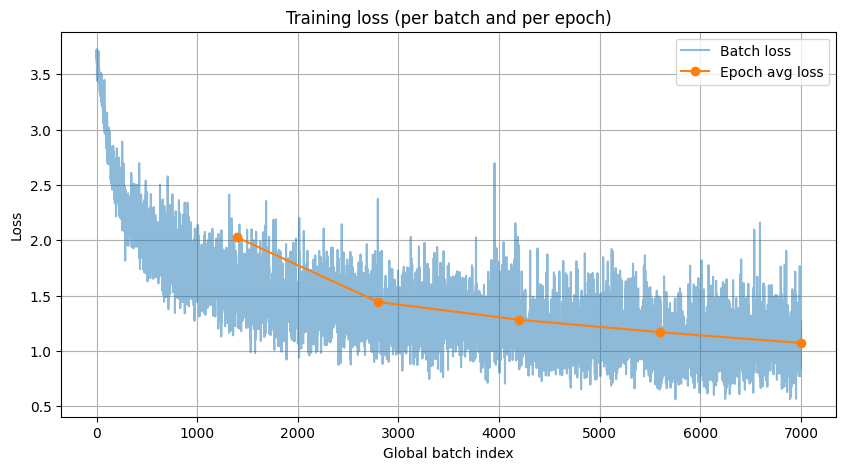}
    \caption{Training loss curve for Dataset~5 (with augmentation).}
    \label{fig:paddy_loss_aug}
\end{figure}

\begin{figure}[h!]
    \centering
    \includegraphics[width=0.5\linewidth]{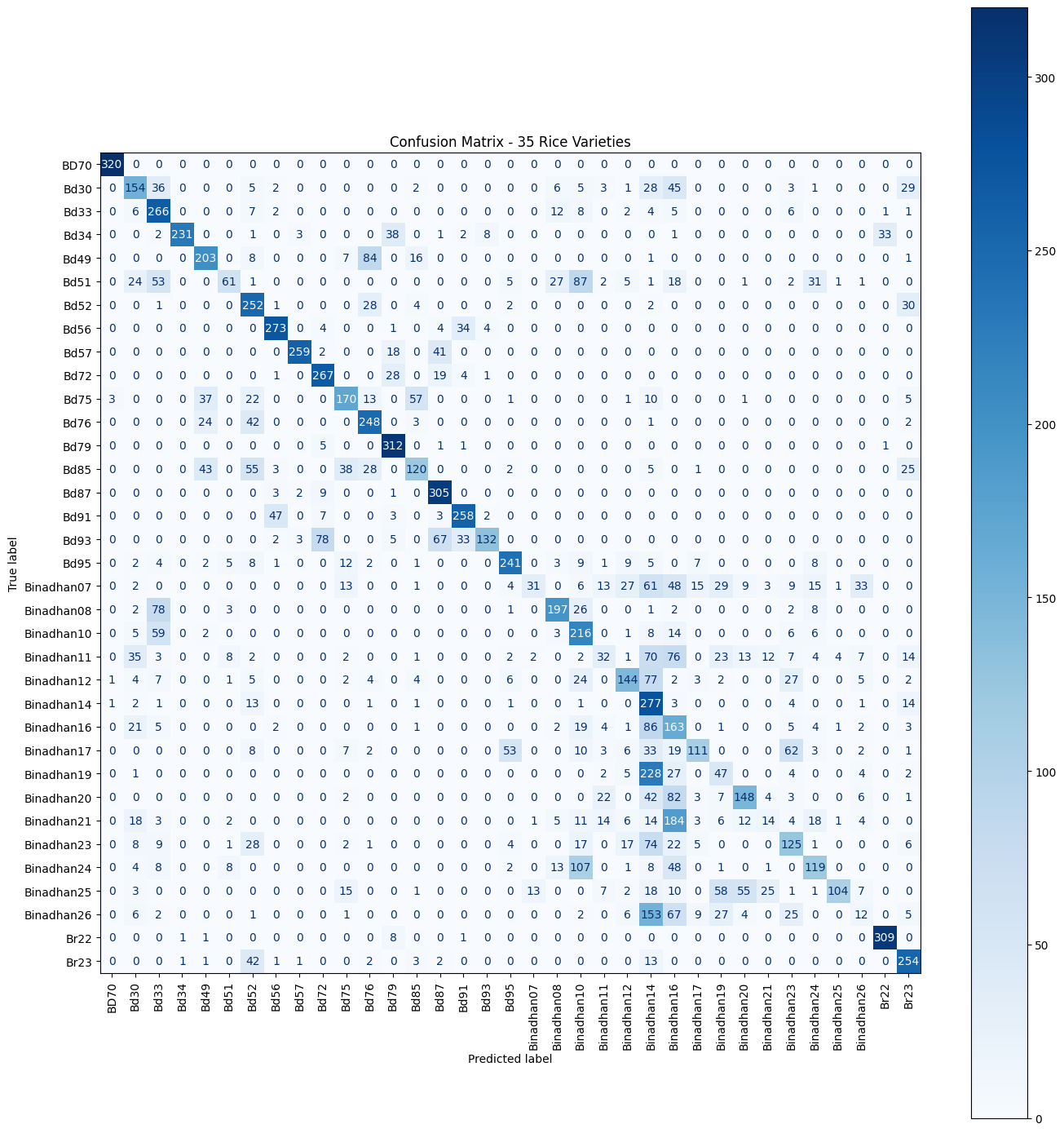}
    \caption{Confusion matrix for Dataset~5 (with augmentation).}
    \label{fig:paddy_cm_aug}
\end{figure}

\subsubsection{Comparative Analysis with Transfer Learning}

Given the complexity of classifying 35 distinct paddy varieties, we benchmarked our model against \textbf{ResNet-50} and \textbf{VGG-16} using Transfer Learning on the augmented dataset. Table~\ref{tab:comparison_dataset5} details the results.

\begin{table}[h!]
\centering
\caption{Comparison on Fine-Grained Dataset 5 (35 Classes, Augmented).}
\label{tab:comparison_dataset5}
\setlength{\tabcolsep}{0pt}
\begin{tabular*}{\textwidth}{@{\extracolsep{\fill}}lccccc}
\hline
\textbf{Model} & \textbf{Method} & \textbf{Accuracy (\%)} & \textbf{F1-Score} & \textbf{Train Time} & \textbf{Params (M)} \\ \hline
\textbf{Custom CNN} & \textbf{Scratch} & \textbf{56.92} & \textbf{0.55} & \textbf{$<$ 15 min} & \textbf{5.41} \\ 
ResNet-50 & Transfer & 86.86 & 0.87 & $\sim$ 45 min & 25.6 \\ 
VGG-16 & Transfer & 86.00 & 0.86 & $\sim$ 40 min & 138.4 \\ \hline
\end{tabular*}
\end{table}

The fine-tuned ResNet-50 achieved the highest accuracy of 86.86\%, followed closely by VGG-16 at 86.00\%. Both heavy architectures significantly outperformed the Custom CNN by a margin of approximately 30\%.


\begin{figure}[H]
    \centering
    \begin{subfigure}[b]{0.32\textwidth}
        \centering
        \includegraphics[width=\linewidth]{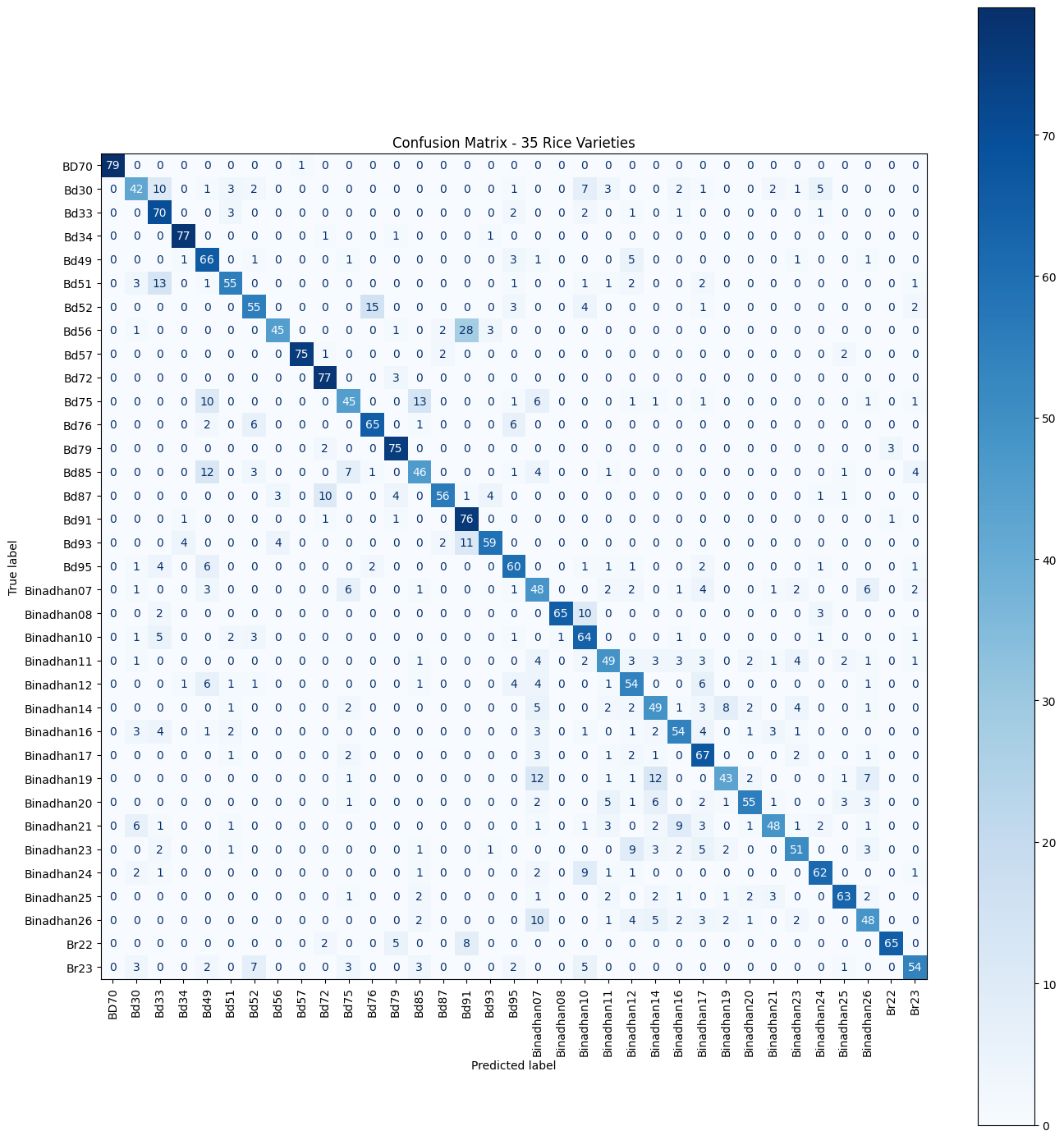}
        \caption{VGG16 (Normal)}
    \end{subfigure}
    \hfill
    \begin{subfigure}[b]{0.32\textwidth}
        \centering
        \includegraphics[width=\linewidth]{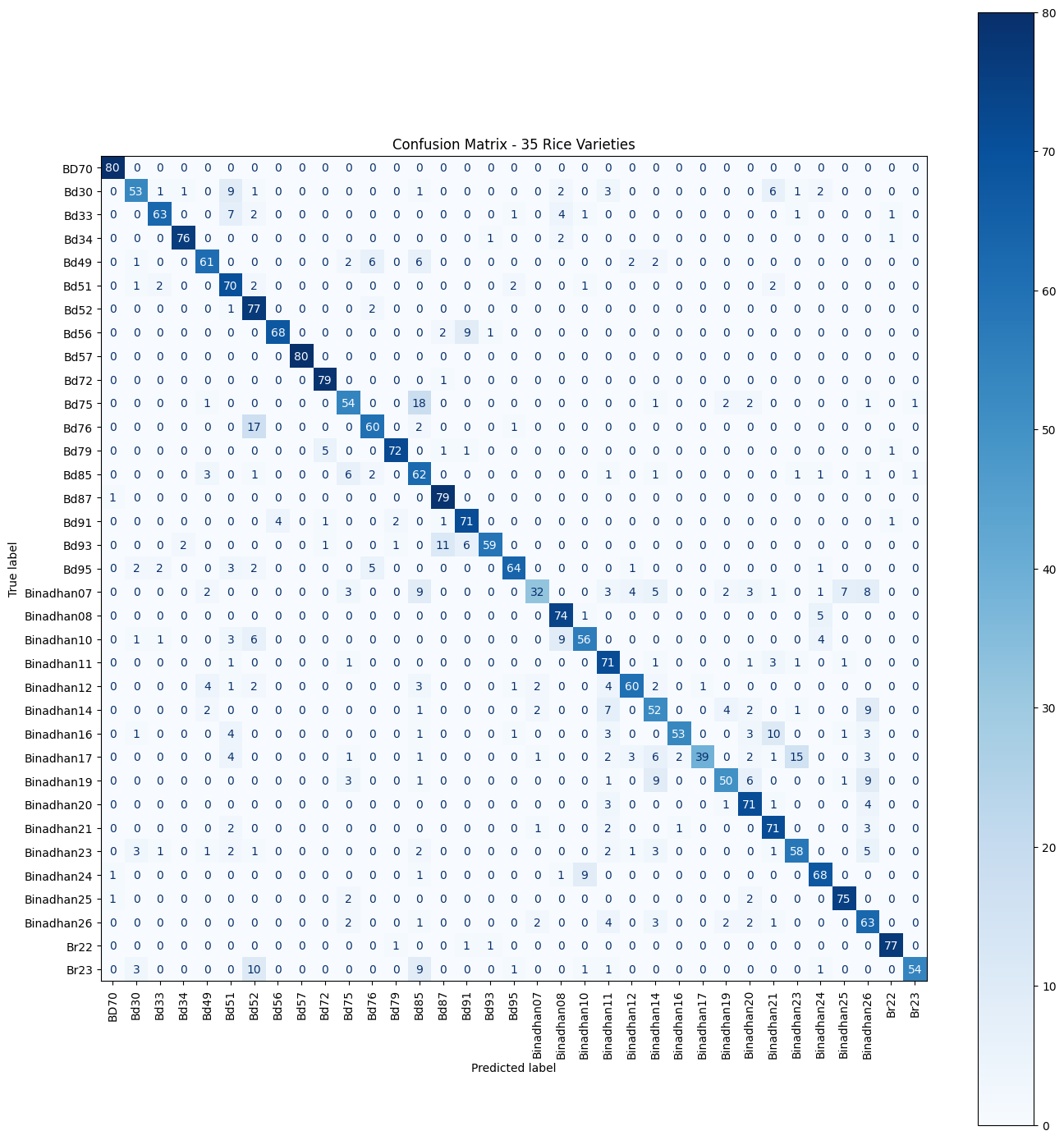}
        \caption{VGG19 (Normal)}
    \end{subfigure}
    \hfill
    \begin{subfigure}[b]{0.32\textwidth}
        \centering
        \includegraphics[width=\linewidth]{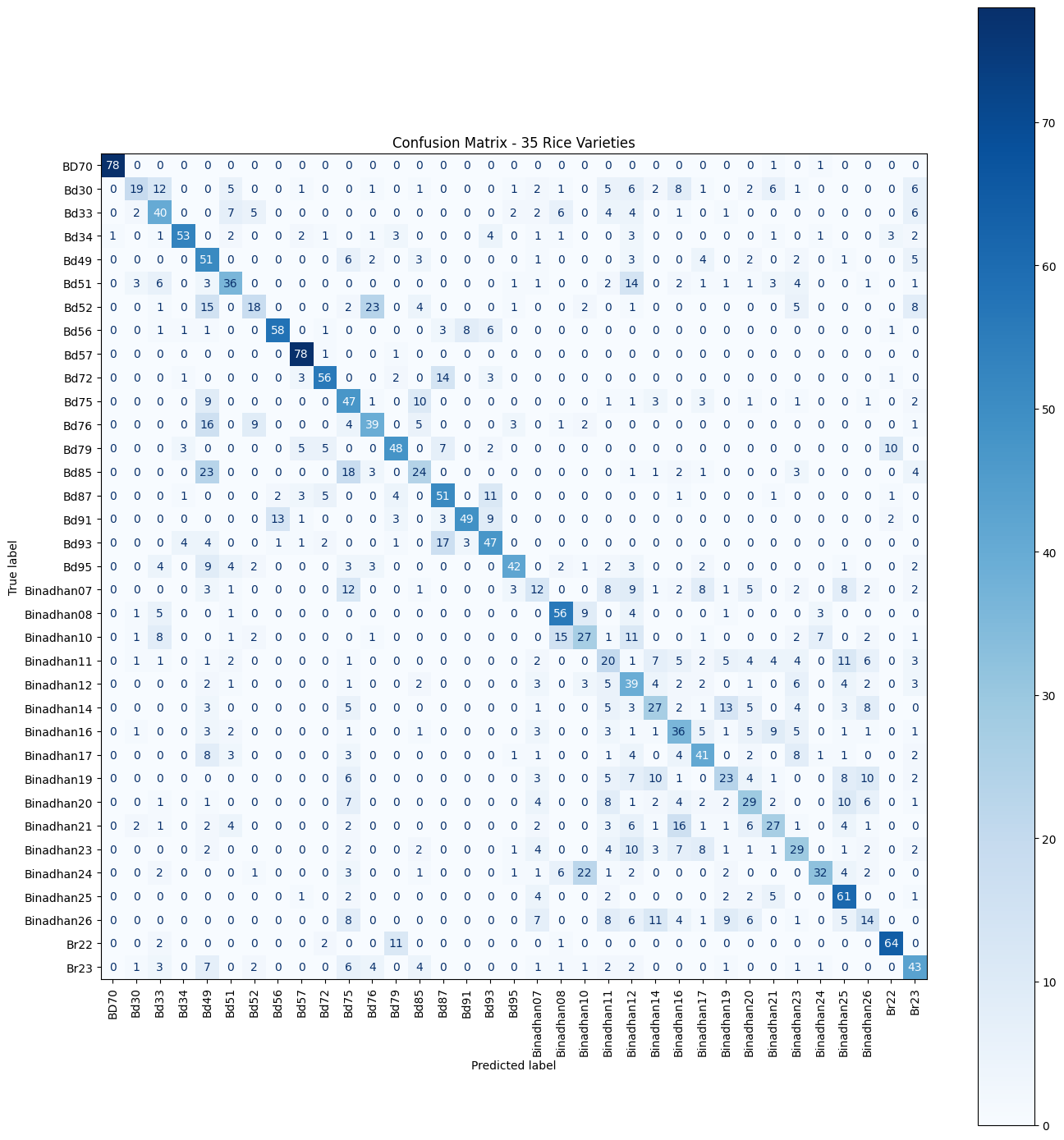}
        \caption{ResNet50 (Normal)}
    \end{subfigure}
    
    \vspace{0.5cm} 
    
    \begin{subfigure}[b]{0.32\textwidth}
        \centering
        \includegraphics[width=\linewidth]{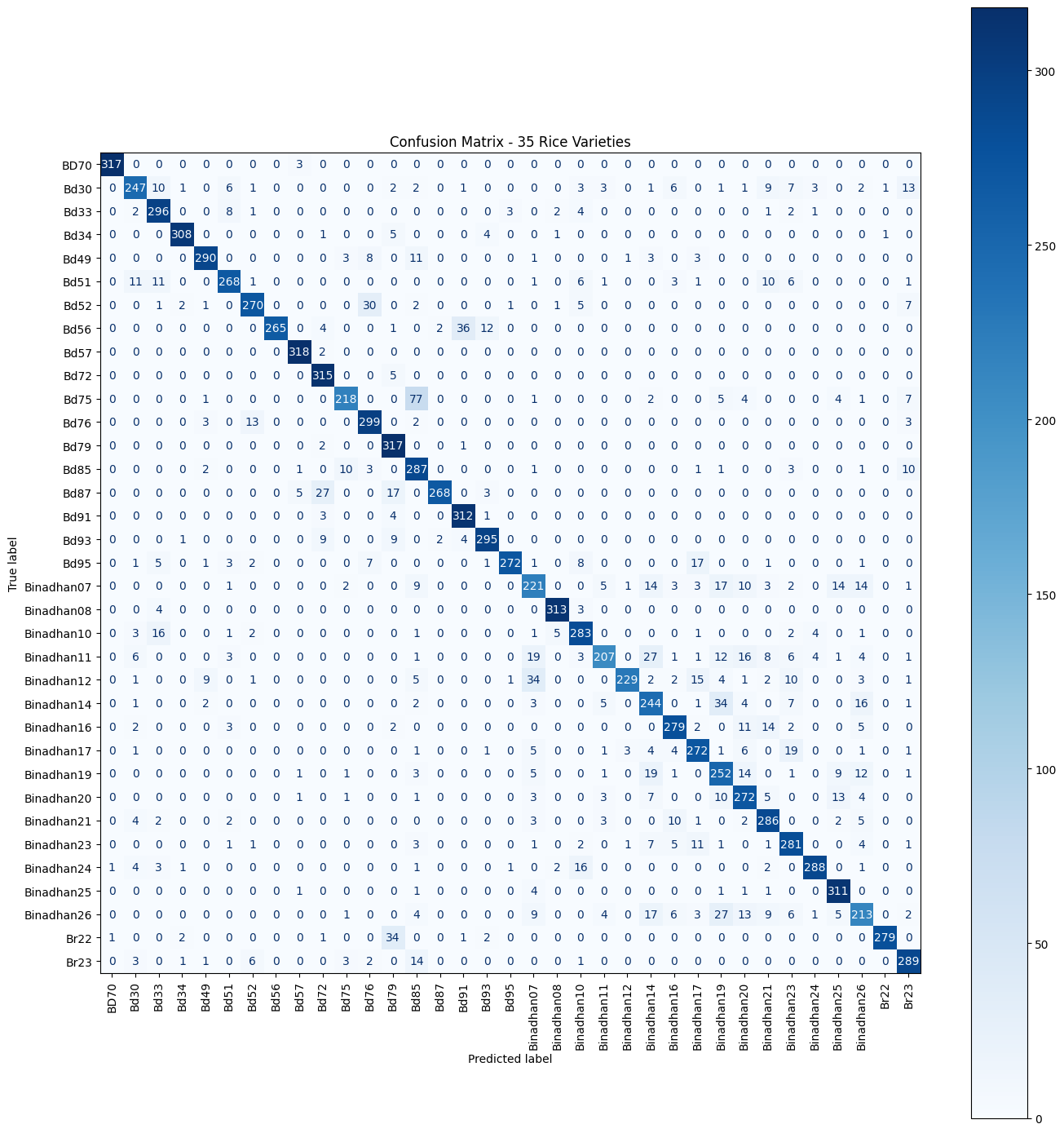}
        \caption{VGG16 (Augmented)}
    \end{subfigure}
    \hfill
    \begin{subfigure}[b]{0.32\textwidth}
        \centering
        \includegraphics[width=\linewidth]{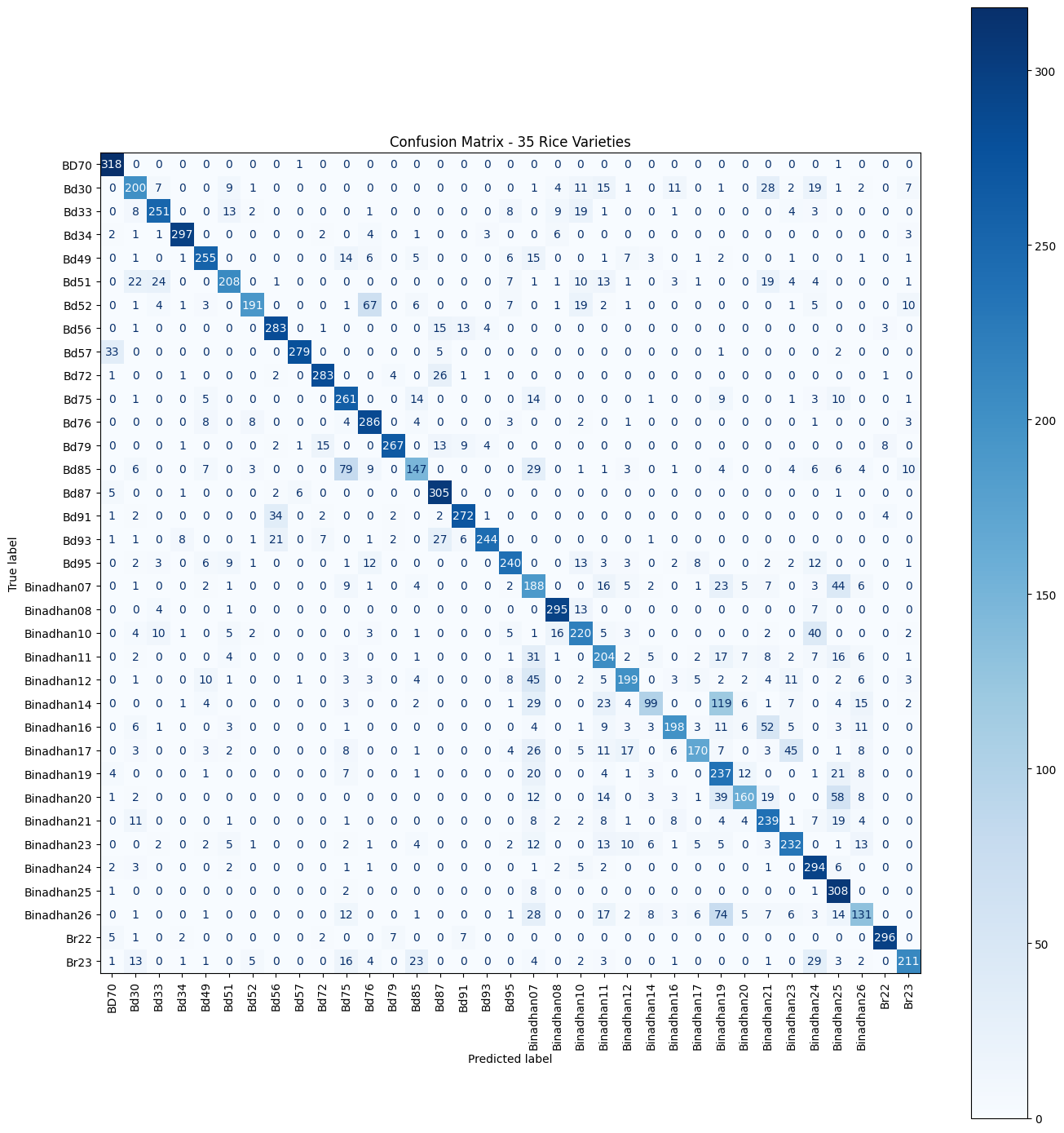}
        \caption{VGG19 (Augmented)}
    \end{subfigure}
    \hfill
    \begin{subfigure}[b]{0.32\textwidth}
        \centering
        \includegraphics[width=\linewidth]{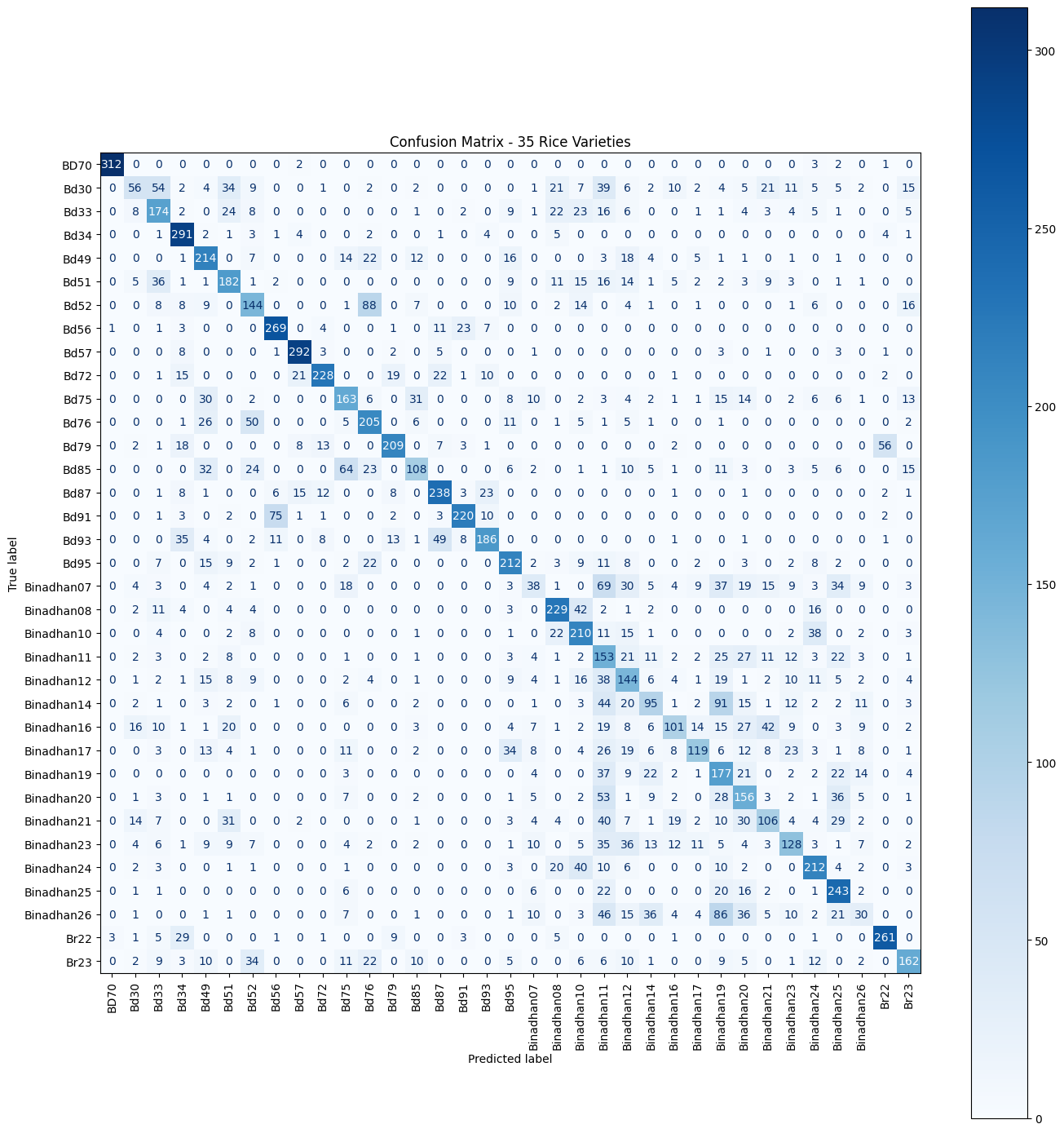}
        \caption{ResNet50 (Augmented)}
    \end{subfigure}
    
    \caption{Confusion Matrix comparison for Dataset 5. The top row displays results from models trained on the normal dataset, while the bottom row displays results using data augmentation. Columns correspond to VGG16, VGG19, and ResNet50 architectures respectively.}
    \label{fig:dataset5_confusion_matrices}
\end{figure}

The results on Dataset 4 \& 5 highlight a critical boundary in the applicability of lightweight models:

\begin{itemize}
    \item \textbf{Fine-Grained Complexity:} Unlike the previous datasets (which were binary or had few distinct classes), this task involves 35 classes with subtle inter-class morphological similarities. The lightweight Custom CNN (5.41M parameters) hits a \textit{capacity bottleneck}, lacking the depth required to disentangle such complex feature spaces.
    \item \textbf{The Necessity of Pre-training:} The superior performance of ResNet-50 and VGG-16 confirms that for high-cardinality tasks, the rich, pre-learned feature representations from ImageNet are invaluable. Learning these features from scratch on a medium-sized dataset is challenging for any architecture, but particularly for a compact one.
    \item \textbf{Computational Cost:} While the Transfer Learning models achieved superior accuracy, they required 3--4$\times$ longer training times (40--45 minutes) compared to the Custom CNN. 
\end{itemize}

In conclusion, for complex, fine-grained classification tasks, the trade-off shifts in favor of Transfer Learning. While the Custom CNN remains efficient, the significant accuracy gap suggests that heavier architectures are justifiable when discerning between a large number of visually similar categories.
\subsection{Overall Discussion Across Datasets}

The comprehensive evaluation across five distinct datasets—ranging from urban surveillance (Datasets 1, 2, 4) to infrastructure assessment (Dataset 3) and agricultural analysis (Dataset 5)—demonstrates the robust adaptability of the proposed lightweight architecture. The experimental results reveal a clear dichotomy based on task complexity:

\begin{itemize}
    \item \textbf{Low-Cardinality Tasks (Datasets 1, 2, 3):} For applications involving a small number of classes (binary or tertiary classification), the proposed custom CNN demonstrated performance parity with, and occasionally surpassed, heavy architectures like ResNet-50. In these scenarios, the massive parameter space of standard deep models often led to overfitting or necessitated extensive regularization. Our compact model, by contrast, learned sufficient discriminative features without the computational overhead, proving that deep transfer learning is often unnecessary for domain-specific, low-complexity tasks.
    
    \item \textbf{High-Cardinality Tasks (Dataset 5):} The limitations of a lightweight feature extractor became more apparent in the fine-grained, 35-class classification task. Here, the semantic richness embedded in pre-trained models provided a distinct accuracy advantage. However, the substantial performance boost observed in our model after data augmentation (increasing accuracy from $\sim$35\% to $\sim$57\%) suggests that the architecture is not inherently incapable; rather, it is highly responsive to data diversity.
\end{itemize}

Overall, the results validate the \textit{efficiency-utility} trade-off: while pre-trained giants offer a "safe bet" for complex tasks, the proposed lightweight CNN offers a superior \textbf{accuracy-per-parameter} ratio. It eliminates the computational debt associated with deploying massive models for simple inference problems, making it the optimal candidate for real-time, edge-based deployment in resource-constrained environments.

\section{Conclusion}\label{sec13}

In this work, we presented a rigorous comparative analysis of three distinct training strategies—custom CNN design, training standard architectures from scratch, and transfer learning—across five diverse real-world datasets. The experimental results demonstrate that while pre-trained models and transfer learning strategies consistently provide a robust baseline, particularly for complex, fine-grained classification tasks, they often incur a computational cost that is disproportionate to the task requirements.

Our proposed lightweight CNN architecture demonstrated remarkable competitiveness, matching or even exceeding the performance of significantly deeper architectures (such as ResNet-50 and EfficientNet-B0) on small- to medium-scale datasets. Although the custom model occasionally yielded slightly lower accuracy on highly complex tasks compared to pre-trained networks, the performance gap was marginal. Crucially, this minor trade-off in accuracy is offset by a substantial reduction in computational complexity. The proposed architecture requires orders of magnitude fewer parameters and lower memory usage, resulting in faster inference times and reduced energy consumption.

Ultimately, this study highlights that "bigger is not always better." For resource-constrained environments and edge computing applications where latency and hardware limitations are critical factors, our streamlined CNN offers a superior accuracy-per-parameter ratio. It provides a pragmatic, efficient alternative to heavy, pre-trained models without sacrificing the reliability required for practical deployment.

\backmatter

\bigskip

\nocite{*} 
\bibliography{sn-bibliography}

\end{document}